\documentclass[10pt,twocolumn,letterpaper]{article}

\usepackage{cvpr}
\usepackage{times}
\usepackage{epsfig}
\usepackage{graphicx}
\usepackage{amsmath}
\usepackage{amssymb}
\usepackage{pifont}
\usepackage[dvipsnames]{xcolor}

\newcommand{\ignore}[1]{}

\newcommand{\ba}{\begin{array}}
\newcommand{\ea}{\end{array}}
\newcommand{\bc}{\begin{center}}
\newcommand{\ec}{\end{center}}
\newcommand{\be}{\begin{enumerate}}
\newcommand{\ee}{\end{enumerate}}
\newcommand{\bea}{\begin{eqnarray}}
\newcommand{\eea}{\end{eqnarray}}
\newcommand{\beas}{\begin{eqnarray*}}
\newcommand{\eeas}{\end{eqnarray*}}
\newcommand{\beq}{\begin{equation}}
\newcommand{\eeq}{\end{equation}}
\newcommand{\bfig}{\begin{figure}}
\newcommand{\efig}{\end{figure}}
\newcommand{\bi}{\begin{itemize}}
\newcommand{\ei}{\end{itemize}}
\newcommand{\bpic}{\begin{picture}}
\newcommand{\epic}{\end{picture}}
\newcommand{\btabular}{\begin{tabular}}
\newcommand{\etabular}{\end{tabular}}
\newcommand{\btable}{\begin{table}}
\newcommand{\etable}{\end{table}}

\newcommand{\es}{\vfill
                 \rule[-6mm]{170mm}{0.7mm} \\
                 \redw{{\tiny
		  \hfill S-\theslide}}
                 \end{slide}}

\newcommand{\matxx}[1]{{\mathtt #1}}
\newcommand{\vecXX}[1]{{\mathbf {#1}}}
\newcommand{\vecYY}[1]{{\boldsymbol {#1}}}

\newcommand{\argmin}{\operatornamewithlimits{arg\ min}}
\newcommand{\argmax}{\operatornamewithlimits{arg\ max}}

\def \hbar {{\bar{h}}}

\def \vech {{\vecXX{h}}}

\def \vecl {{\vecXX{l}}}

\def \vect {{\vecXX{t}}}

\def \vecv {{\vecXX{v}}}

\def \vecx {{\vecXX{x}}}

\def \vecz {{\vecXX{z}}}

\def \veceta   {{\vecYY{\eta}}}

\def \matJ {{\matxx{J}}}

\def \matSigma  {{\matxx{\Sigma}}}
\def \matLambda {{\matxx{\Lambda}}}

\makeatletter
\renewcommand*\env@matrix[1][*\c@MaxMatrixCols c]{%
  \hskip -\arraycolsep
  \let\@ifnextchar\new@ifnextchar
  \array{#1}}
\makeatother

\usepackage{amsmath,mleftright}
\usepackage{xparse}

\usepackage{subfig}
\usepackage{graphicx}
\usepackage{float}

\NewDocumentCommand{\evalat}{sO{\big}mm}{%
  \IfBooleanTF{#1}
  {\mleft. #3 \mright|_{#4}}
  {#3#2|_{#4}}%
}

\usepackage[pagebackref=true,breaklinks=true,letterpaper=true,colorlinks,bookmarks=false]{hyperref}

\cvprfinalcopy 

\def\cvprPaperID{9635} 

\ifcvprfinal\fi
\begin{document}

\title{Bundle Adjustment on a Graph Processor}

\author{Joseph Ortiz$^{1}$, Mark Pupilli$^{2}$, Stefan Leutenegger$^{1}$, Andrew J. Davison$^{1}$\\
$^{1}$Imperial College London, Department of Computing, UK. $\;$ $^{2}$Graphcore. \\
{\tt\small j.ortiz@imperial.ac.uk}
}

\maketitle

\begin{abstract}
Graph processors such as Graphcore's Intelligence Processing Unit (IPU) are part of the major new wave of novel computer architecture for AI, and have a general design with massively parallel computation, distributed on-chip memory and very high inter-core communication bandwidth which allows breakthrough performance for message passing algorithms on arbitrary graphs.

We show for the first time that the classical computer vision problem of bundle adjustment (BA) can be solved extremely fast on a graph processor using Gaussian Belief Propagation. Our simple but fully parallel implementation uses the 1216 cores on a single IPU chip to, for instance, solve a real BA problem with 125 keyframes and 1919 points in under 40ms, compared to 1450ms for the Ceres CPU library. Further code optimisation will surely increase this difference on static problems, but we argue that the real promise of graph processing is for flexible in-place optimisation of general, dynamically changing factor graphs representing Spatial AI problems. We give indications of this with experiments showing the ability of GBP to efficiently solve incremental SLAM problems, and deal with robust cost functions and different types of factors.
\end{abstract}

\section{Introduction}

Real-world applications which require a general real-time `Spatial AI' capability from computer vision are becoming more prevalent in areas such as robotics, UAVs and AR headsets, but it is clear that a large gap still exists between the ideal performance required and what can be delivered within the constraints of real embodied products, such as low power usage.
An increasingly important direction is the design of processor and sensor hardware specifically for vision and AI workloads to replace the general purpose CPUs, GPUs and frame-based video cameras which are currently prevalent~\cite{Davison:ARXIV2018,Nardi:etal:ICRA2015}. The space of AI and vision algorithm design continues to change rapidly and we believe that it is not the right time to make very specific decisions such as `baking in' a particular SLAM algorithm to processor hardware, except perhaps for very specific use cases.

\bfig
    \centering
    \includegraphics[scale=0.45]{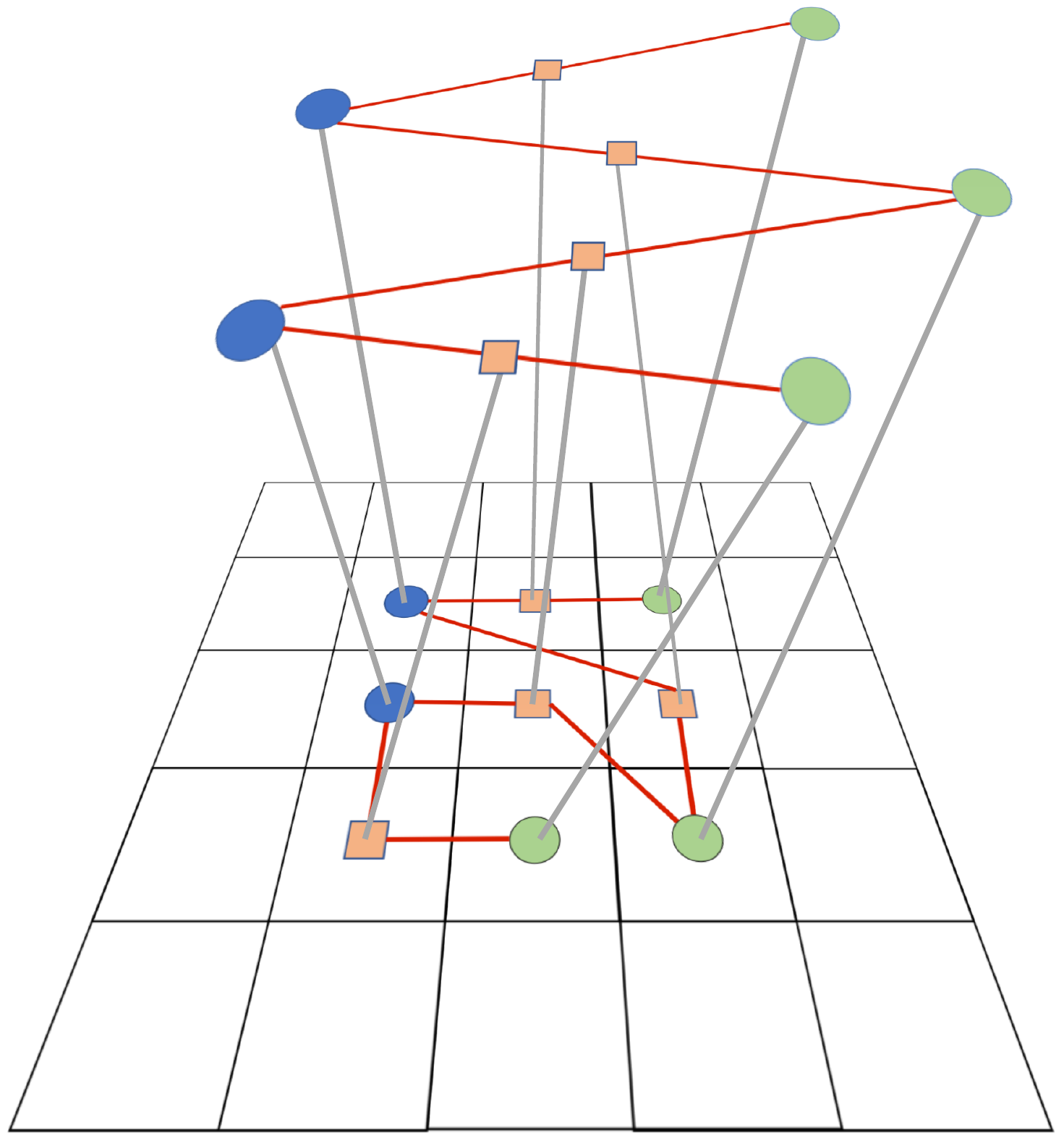}
    \caption{We map a bundle adjustment factor graph onto the tiles (cores) of Graphcore's IPU and show that Gaussian Belief Propagation can be used for rapid, distributed, in-place inference for large problems. Here we display the most simple mapping in which each node in the factor graph is mapped onto a single arbitrary tile. Keyframe nodes are blue, landmark nodes are green and measurement factor nodes are orange. }
    \label{fig:mapping}
    \vspace{2mm} \hrule
\efig

However, new architectures are emerging which have made quite general design choices about processing for AI workloads.
Efficient and low power computation must be massively parallel and minimise data transfer. To this end, storage and processing should be distributed, and as much computation as possible should happen `in place'. A key example is Graphcore's Intelligence Processing Unit (IPU) \cite{Graphcore}, which implements this concept within a single large chip 
which is composed of 1216 cores called \textit{tiles}, each with local memory arranged in a fully connected graph structure. 
It is massively parallel like a GPU, but its tiles have a completely different interconnect structure.
The IPU has breakthrough performance for algorithms which have a sparse graph message passing character.
The key early commercial use case for the IPU is as a flexible deep learning accelerator \cite{Lacey:IPUBenchmarks2019}, primarily in the cloud, but we believe that it has much more general potential for Spatial AI computation.

In this paper we consider bundle adjustment (BA), a central element of 3D visual processing which is representative of many geometric estimation problems, and show that Gaussian Belief Propagation (GBP) can perform rapid optimisation of BA problems on a single IPU chip.

GBP is a special case of general loopy belief propagation, a well known technique in probabilistic estimation, but it has previously only been minimally used in geometric vision and robotics problems~\cite{Davison:Ortiz:ARXIV2019}. It is an algorithm which can be run on a CPU, but is not necessarily competitive there compared to  alternative optimisation techniques which take global account of the structure of a problem. However, GBP can be mapped to a graph processor due to its fully distributed nature to take full advantage of the massively parallel capability of an IPU.

We present the first implementation of BA on a graph processor, with breakthrough optimisation speed for a variety of diverse sequences in which we record an average speed advantage 24x over the Ceres library on a CPU. Our implementation is simple and preliminary, implemented with only 1000 lines of Poplar\texttrademark C++ code, and there is surely much room for future performance optimisation.

Positive characteristics of our GBP approach include:
extremely fast local convergence, the ability to use robust cost functions to reject outlying measurements, and the ability to easily deal with dynamic addition of variables and data and rapidly re-optimise solutions. We highlight these aspects in our results, and argue as in~\cite{Davison:Ortiz:ARXIV2019}
for the huge potential for graph processing and GBP in general incremental factor graph optimisation for Spatial AI. It would be straightforward and efficient  to incorporate factors from additional 
priors and sensors into this framework, such as smoothness of scene regions due to recognition, and continue to optimise for global estimates with all computation and storage done in-place on a graph processor.

\section{Related Work}

Factor graphs are commonly used in geometric vision to represent the structure of constraints in estimation problems \cite{Bloesch:etal:CVPR2018, Engel:etal:PAMI2017, Folkesson:Christensen:ICRA2004, Kaess:etal:TRO2008, Lu:Milios:AR1997, Mur-Artal:etal:TRO2015}. In particular, for bundle adjustment \cite{Triggs:etal:VISALG1999} researchers have leveraged the global structure of these constraints to design efficient inference algorithms \cite{Agarwal:etal:ICCV2009, Jeong:etal:CVPR2010}.

Several works have taken the approach of converting the loopy factor graph into a tree \cite{Kaess:etal:IJRR2012, Paskin:IJCAI2003}. 
iSAM2 \cite{Kaess:etal:IJRR2012} uses variable elimination to convert the loopy factor graph to a Bayes tree while \cite{Paskin:IJCAI2003} uses a junction tree-like method which employs maximum likelihood projections to remove edges. This category of methods differs from our approach in that it requires periodic centralised computation to convert the loopy constraint graph into a tree. 

More closely related to our work, \cite{Crandall:etal:CVPR2011} and \cite{Ranganathan:etal:IJCAI2007} use Loopy Belief Propagation for geometric estimation problems, though with CPU implementation. \cite{Crandall:etal:CVPR2011} uses discrete BP to provide an initialisation for Levenberg-Marquardt refinement in BA, and Loopy SAM \cite{Ranganathan:etal:IJCAI2007} uses GBP to solve a SLAM-like problem for a relatively small 2D scene. 

In the domain of computer architecture, there has been substantial recent effort to design specific hardware for vision algorithms \cite{Saeedi:etal:2018, Zhang:etal:RSS2017}. This is particularly evident in industry, where we have seen development of chips such as the HoloLens' HPU and the Movidius VPU series, though the main accelerations achieved to date have been in vision front-ends such as feature matching.

Other related research has made use of parallelism on existing hardware to accelerate BA. Multicore BA \cite{Wu:etal:BA:CVPR2011} proposed an inexact but parallelisable implementation for CPUs or GPUs, while \cite{Gupta:etal:ECCV2010} advocated a hybrid GPU and CPU implementation. More generally, \cite{DeVito:etal:TOG2017} accelerated non-linear least squares problems in graphics by automatically generating GPU solvers.

\section{Preliminaries}

\subsection{Factor Graphs}

Factor graphs are well known in geometric vision as a representation of the structure of estimation problems.
A factor graph, $G=(V, F, E)$, is a bipartite graph composed of a set of variable nodes $V=\{\vecv_i\}_{i=1:N_v}$, a set of factor nodes $F=\{f_s\}_{s=1:N_f}$ and a set of edges $E$. Each factor node $f_s$ represents a probabilistic constraint between a subset of variables $V_s \subset V$ which is described by an arbitrary function $f_s(V_s)$. The factorisation is explicitly represented in the graph by connecting factor nodes with the variable nodes they depend on. Probabilistically speaking, these factors are the independent terms that make up the joint distribution:
\beq
\label{eqn:factorprod}
p(V) = \prod_{s=1}^{N_f} f_s(V_s)
~.
\eeq

\subsection{Belief Propagation \label{sec:bp}}

Belief propagation (BP) \cite{Pearl:book1988} is a well-known distributed inference algorithm for computing the marginal distribution for a set of variables from their joint distribution. The marginal for a single variable $\vecv_i$ is the integral of the joint distribution over all other variables: 
\beq
\label{eqn:marginal}
    p(\vecv_i) = \int p(V) \;d\vecv_1 \,...\, d\vecv_{i-1} d\vecv_{i+1} \,...\, d\vecv_{N_v}
    ~.
\eeq
BP works by passing messages through the factor graph and is efficient as it leverages the fact that the topology of the graph encodes the factorisation of the joint distribution. The marginals are computed using iterative local message passing which alternates between factor nodes sending messages to variable nodes and variable nodes sending messages to factor nodes. See \cite{Bishop:Book2006} or  \cite{Davison:Ortiz:ARXIV2019} for a derivation of the message passing rules.

By design, belief propagation infers the marginals for tree graphs in one sweep of messages from the root node to the leaf nodes and then back up. For loopy graphs, the same BP message passing can be applied with a message passing schedule, and after many iterations estimates converge to the marginals. Loopy BP does not have convergence guarantees, however it is generally stable \cite{Murphy:etal:1999}. When the distributions are represented as Gaussians, Loopy Gaussian Belief Propagation converges to the correct marginal posterior means for all graph topologies
\cite{Weiss:Freeman:NIPS2000}.

Key to understanding why belief propagation is efficient is considering the least efficient way to compute the marginal distribution for a variable. The naive way would be to take a product of all of the factors to give the joint distribution and then marginalise over all other variables. This simultaneous marginalisation over all other variables is expensive; for example, in the discrete case, if each variable takes $k$ discrete values then marginalising over all but one variable requires summing $k^{N_v - 1}$ terms. 
Belief propagation instead marginalises over minimal independent subsets of variables using the conditional dependency information which is encoded in the graph topology. 
Returning to the example of discrete variables, if we want to compute the marginal distribution for a tree graph containing only pairwise factors, belief propagation requires summing only $2  N_f k^2$ terms.

\section{The Bundle Adjustment Factor Graph}

\bfig
\centering
    \includegraphics[scale=0.38]{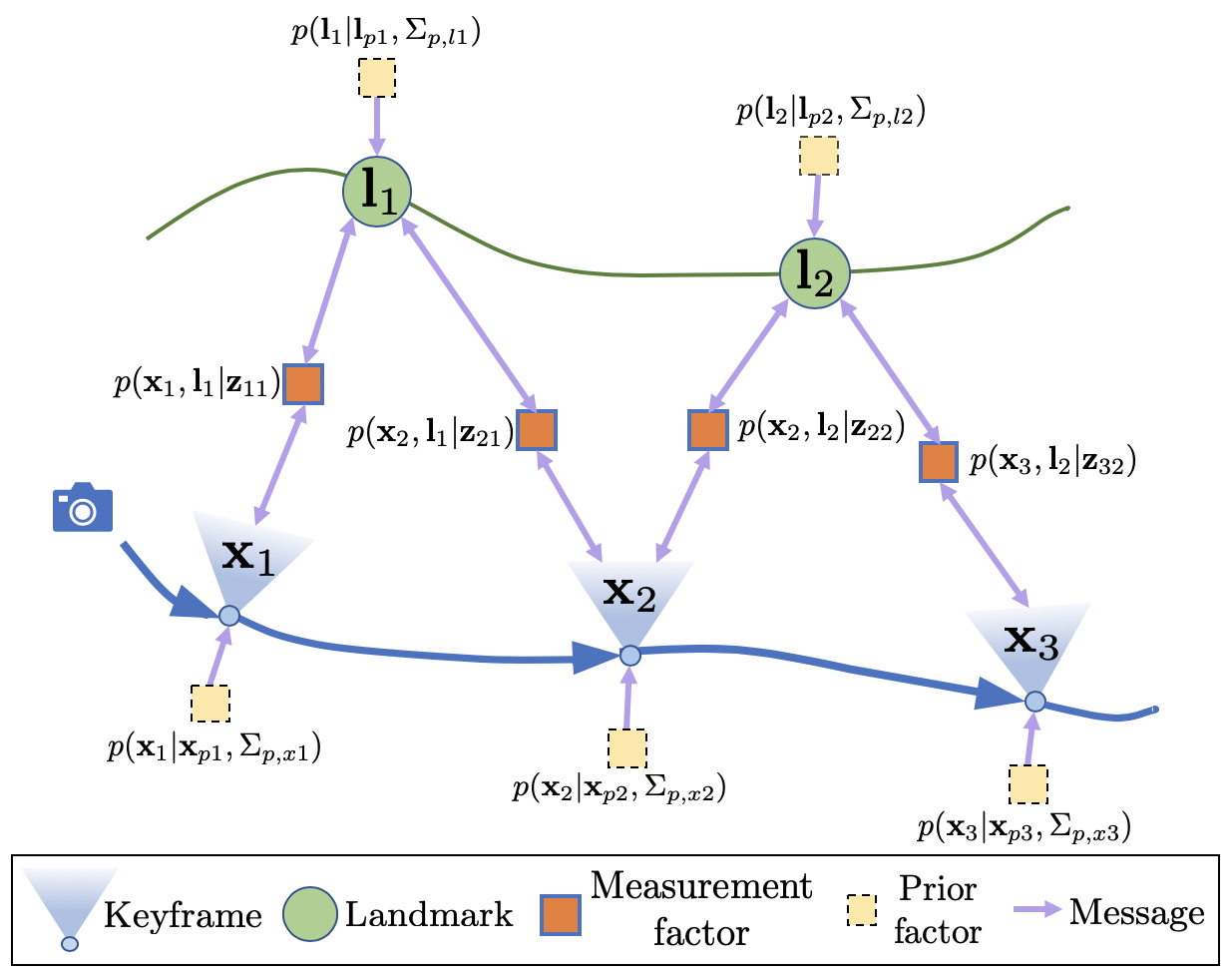}
    \caption{\textbf{Factor graph illustration.} Measurement factors connect keyframes and the landmarks they observe. Keyframes and landmarks are instantiated with an automatically generated weak prior factor. Messages are sent from all factors to adjacent keyframe and landmark nodes and from keyframe and landmark nodes to adjacent measurement factor nodes.}
    \label{fig:factor_graph}
    \vspace{2mm} \hrule
\efig

Bundle adjustment is the problem of jointly refining the set of variables $V=X \cup L$, where $X = \{ \vecx_i \}_{i=1:N_k}$ is the set of keyframe poses and $L = \{ \vecl_j \}_{j=1:N_l}$ is the set of landmark locations, subject to a set of constraints which define the error we want to minimise. Specifically, we include two types of error terms: reprojection errors and prior errors. The reprojection error penalises the distances between the projections of landmarks into the image plane of the keyframes that observe them and the set of measurements corresponding to these observations $Z = \{ \vecz_{km} \}$. The prior error terms try to maximise the probability that the current variable values were drawn from the corresponding prior distribution $\{ \mathcal{N}(\vecx_i ; \vecx_{p_i}, \matSigma_{p,xi}), \; \mathcal{N}( \vecl_j ; \vecl_{p_j}, \matSigma_{p,lj}) \}_{i=1:N_k, j=1:N_l}$.  The prior terms are required to set the overall scale for monocular problems and to condition the messages from the measurement factors which would otherwise only constrain 2 degrees of freedom. Given an initialisation point, the priors are automatically generated such that they are a factor of 100 weaker than the reprojection error terms in the objective. We formulate this using the Jacobians and the measurement model which define the strength of measurement constraints. An example factor graph for a small BA problem is shown in Figure \ref{fig:factor_graph}. 

In bundle adjustment we want to perform maximum a posteriori (MAP) inference which computes the configuration of variables $\{X, L\}$ that maximises the joint probability $p(X, L | Z)$:
\bea
    \label{eqn:maxjoint}
    \{X^*, L^*\} &=& \argmax_{\{X, L\}} p(X, L | Z) \\
    &=& \argmax_{\{X, L\}} p(Z| X, L) p(X, L) 
    ~.
\eea

\noindent In the second line we have used Bayes theorem and dropped the denominator $p(Z)$ as measurements are given quantities and do not affect the MAP solution. This leads to the factorisation of the probability distribution that we want to maximise (which we will call $p_{\mathrm{obj}}(X, L)$) into the product of the likelihood of the measurements given the variables $p(Z| X, L)$ and priors on the variables $p(X, L)$. As $\vecx_i$ and $\vecx_j$ are independent in our formulation, $\vecl_i$ and $\vecl_j$ are independent and $\vecx_i$ and $\vecl_j$ are only conditionally dependent given a measurement $\vecz_{ij}$, these terms can be further factorised:
\beq
\label{eqn:jointprob}
    \small
    p_{\mathrm{obj}}(X, L) = \prod_{i=1}^{N_k} \phi_i(\vecx_i)  \prod_{j=1}^{N_l} \theta_j(\vecl_j) \prod_{k=1}^{N_k} \, \prod_{m, \vecl_m \in L_k}   \psi_{km}(\vecx_k, \vecl_m) 
    ~,
\eeq

\noindent where $L_k$ is the set of landmarks observed by keyframe $\vecx_k$.

The set of factors $\{\phi_i, \theta_j, \psi_{km} \}_{i=1:N_k, j=1:N_l, {km \in O}}$ can be interpreted as prior constraints on the keyframe poses, prior constraints on the landmark positions and measurement reprojection constraints respectively. The prior constraints have the form of Gaussians over the variables $\{ \vecx_i \}_{i=1:N_k}$ and $\{ \vecl_j \}_{j=1:N_l}$:
\bea
    \small
    \phi_i(\vecx_i) &=& p(\vecx_i | \vecx_{p_i}, \matSigma_{p,xi}) \\
    &\propto& \exp{(-\frac{1}{2} \parallel \vecx_i - \vecx_{p, i} \parallel^2_{\matSigma_{p,xi}} )}
    ~,
    \label{eqn:camprior}
\eea

\vspace{-7mm}

\bea
    \small
    \theta_j(\vecl_j) &=& p(\vecl_j | \vecl_{p_j}, \matSigma_{p,lj}) \\
    &\propto& \exp{(-\frac{1}{2} \parallel \vecl_j - \vecl_{p, j} \parallel^2_{\matSigma_{p,lj}} )}
    ~.
    \label{eqn:lmkprior}
\eea

Assuming a Gaussian measurement model, $\vecz_{km} = \vech(\vecx_k,\vecl_m) + \eta$, with $\eta \sim \mathcal{N}(0, \matSigma_M)$ we can write out the form of the measurement factors:
\beq
    \psi_{km}(\vecx_k, \vecl_m) = p(\vecx_k, \vecl_m| \vecz_{km}) \propto p(\vecz_{km}|\vecx_k, \vecl_m)  
\eeq
\vspace{-2mm}
\beq
    \propto \exp{(-\frac{1}{2} \parallel \vecz_{km} - \vech(\vecx_k,\vecl_m) \parallel^2_{\matSigma_M} ) }
    ~.
    \label{eqn:meas_constraint}
\eeq


The measurement factor $\psi_{km}$ is Gaussian in $\vecz_{km}$ but is Gaussian in the variables $\vecx_k$ and $\vecl_m$ only if the measurement function $\vech(\vecx_k,\vecl_m)$ is linear. In our case, we have a nonlinear measurement function, $\vech(\vecx_k,\vecl_m) = \pi \, (R_k \, \vecl_m + \vect_k)$, where $\pi$ is the projection operator and $R_k$ and $\vect_k$ are the rotations and translations derived from $\vecx_k$ . 
As a result, we must update the measurement factors by relinearising during optimisation. 

After linearising about some fixed point $(\vecx_{k, 0}, \vecl_{m, 0})$, the measurement factors can be expressed as a Gaussian distribution using the information form which is parametrised by an information vector $\veceta$ and information matrix $\matLambda$:
\beq
    \label{eqn:infform}
    \mathcal{N}^{-1}(\vecx;\veceta, \matLambda) \propto \; \exp{(-\frac{1}{2} \vecx^\top \matLambda \vecx + \veceta^\top \vecx)}
    ~.
\eeq
\noindent The information form is used as it can represent distributions with rank deficient covariances in which a variable is not constrained at all along a particular direction. With this at hand and after a small amount of work \cite{Davison:Ortiz:ARXIV2019}, we find that linearised measurement factors take the following form:

\beq
    \psi_{km}(\vecx_k, \vecl_m) = \mathcal{N}^{-1}\bigg(\begin{bmatrix} \vecx_k \\ \vecl_m \end{bmatrix};\veceta_{km}, \matLambda_{km}\bigg)
    ~ ,
\eeq
\noindent where,
\beq
    \veceta_{km} =  \matJ^\top \matSigma_M^{-1} \bigg( \matJ\; \begin{bmatrix} \vecx_{k, 0} \\ \vecl_{m, 0} \end{bmatrix} + \vecz_{km} - \vech(\vecx_{k, 0}, \vecl_{m, 0}) \bigg) 
    ~ ,
\eeq
\beq
    \matLambda_{km} =  \matJ^\top \matSigma_M^{-1} \matJ 
    ~ , 
\eeq

\noindent and the $2\times 9$ Jacobian $\matJ =  \evalat[\big] {\begin{bmatrix} \frac{\partial \vech}{\partial \vecx_{k}} , \frac{\partial \vech}{\partial \vecl_{m}}  \end{bmatrix}} {\vecx_{k} = \vecx_{k, 0}, \vecl_{m} = \vecl_{m, 0} }$.

Now that all of our constraints are in the Gaussian form, finding the MAP solution is equivalent to minimising the negative log likelihood which is a sum of squared residuals:
\bea
\label{eqn:residuals}
\begin{aligned}
    \{X^*, L^*\} = \argmin_{\{X, L\}} \: \bigg[ \;\sum_{i=1}^{N_k} \parallel \vecx_i - \vecx_{p, i} \parallel^2_{\matSigma_{p,xi}} + \\
    \sum_{j=1}^{N_l} \parallel \vecl_j - \vecl_{p, j} \parallel^2_{\matSigma_{p,lj}} + \\
    \sum_{k=1}^{N_k} \sum_{m, \vecl_m \in L_k} \parallel \vecz_{km} - \vech(\vecx_k,\vecl_m) \parallel^2_{\matSigma_M} \:\bigg] .
\end{aligned}
\eea

\section{Gaussian Belief Propagation for Bundle Adjustment}

GBP is a Bayesian algorithm that can be used to solve bundle adjustment problems by computing the marginal distribution, with mean equal to the MAP solution, for all variables. In contrast, classical bundle adjustment methods compute a point estimate of the MAP solution using the Levenberg-Marquardt algorithm. 

As the bundle adjustment factor graph is loopy, GBP stores a belief distribution at each variable node which converges to the marginal distribution after sufficient iterations of message passing. To describe the message passing equations, we do not distinguish between keyframe and landmark variable nodes and denote a variable node from the set $V = X \cup L$ as $\vecv_i$ and the belief stored at this node at iteration t, $b_i^t(\vecv_i) = \mathcal{N}^{-1}(\vecv_i; \veceta_{b_i}^t, \matLambda_{b_i}^t)$. 

Prior factors send the same message, $pr_i(\vecv_i) = \mathcal{N}^{-1}(\vecv_i; \veceta_{p_i}, \matLambda_{p_i})$, to the variable node they connect to at all iterations. To describe the messages from measurement factors, we must first divide up the parameters of the factor distribution: 
\beq
\label{eqn:factor}
    \psi_{ij}\bigg(\begin{bmatrix} \vecv_i \\ \vecv_j \end{bmatrix}\bigg) = \mathcal{N}^{-1}\bigg(\begin{bmatrix} \vecv_i \\ \vecv_j \end{bmatrix}; \begin{bmatrix} \veceta_i^{ij} \\ \veceta_j^{ij} \end{bmatrix}, \begin{bmatrix} \matLambda_{ii}^{ij} &&   \matLambda_{ij}^{ij} \\ \matLambda_{ji}^{ij} &&   \matLambda_{jj}^{ij} \end{bmatrix} \bigg) \\
    ~.
\eeq

\noindent The message passing rules \cite{Bishop:Book2006} dictate that a pairwise factor $\psi_{ij}$ computes the message to variable node $\vecv_i$ by taking  the product of its factor distribution and the message from variable node $\vecv_j$ before marginalising over $\vecv_j$. After this calculation, the message from measurement factor $\psi_{ij}$ to $\vecv_i$ at iteration $t + 1$, $\mu^{t+1}_{j \rightarrow i}(\vecv_i) = \mathcal{N}^{-1}(\vecv_i; \veceta^{t+1}_{j \rightarrow i}, \matLambda^{t+1}_{j \rightarrow i})$, has the form :
\beq
\label{eqn:messageeta}
    \small
    \veceta^{t+1}_{j \rightarrow i} = \veceta_i^{ij} - \matLambda_{ij}^{ij} \, (\matLambda_{jj}^{ij} +  \matLambda^t_{b_j} - \matLambda^{t}_{i \rightarrow j})^{-1} (\veceta_j^{ij} + \veceta^{t}_{b_j} - \veceta^{t}_{i \rightarrow j})
    ~,
\eeq
\beq
\label{eqn:messagelambda}
    \matLambda^{t+1}_{j \rightarrow i} = \matLambda_{ii}^{ij} - \matLambda_{ij}^{ij} \, (\matLambda_{jj}^{ij} +  \matLambda^t_{b_j} - \matLambda^{t}_{i \rightarrow j})^{-1} \matLambda_{ji}^{ij}
    ~.
\eeq

Variable nodes update their belief by taking a product of incoming messages from their prior factor and all adjacent measurement factors. The belief information vector and information matrix are updated as follows:

\beq
\label{eqn:beliefeta}
    \veceta^{t+1}_{b_i} = \veceta_{p_i} + \sum_{ j, \psi_{ij} \in n(\vecv_i)} \veceta^t_{j \rightarrow i}
    ~,
\eeq

\beq
\label{eqn:belieflambda}
    \matLambda^{t+1}_{b_i} = \matLambda_{p_i} + \sum_{j, \psi_{ij} \in n(\vecv_i)} \matLambda^t_{j \rightarrow i}
    ~,
\eeq

\noindent where the function $n(.)$ returns the adjacent nodes. The beliefs are sent as messages from the variable nodes to the factor nodes as the true message can be recovered at the factor node using the previous factor to variable message. 

We use a synchronous scheduling, in which, at each iteration, all factor nodes relinearise and send messages to adjacent variable nodes before all variable nodes update their belief and send back messages to adjacent factor nodes. In our framework, relinearisation is done in an entirely local manner and a measurement factor is relinearised when the distance between the current belief estimate and the linearisation point of the variables the factor connects to is greater than a threshold $\beta$.

After sufficient iterations of message passing and relinearisation, the belief distributions converge to the marginal distributions:
\beq
\label{eqn:converge}
    b^t_i(\vecv_i) \rightarrow p(\vecv_i)
    ~.
\eeq
A final detail to note is that we use message damping which is commonly used to stabilise the convergence of Loopy GBP \cite{Malioutov:etal:2006}. We damp the update in Equation \ref{eqn:messageeta}, such that $\veceta^{t+1}_{j \rightarrow i}$ is replaced with $(1-d) \, \veceta^{t+1}_{j \rightarrow i} + d \, \veceta^{t}_{j \rightarrow i}$, where $d$ is a damping factor. 

\begin{figure*}[h]
    \centering
    \includegraphics[width=\linewidth]{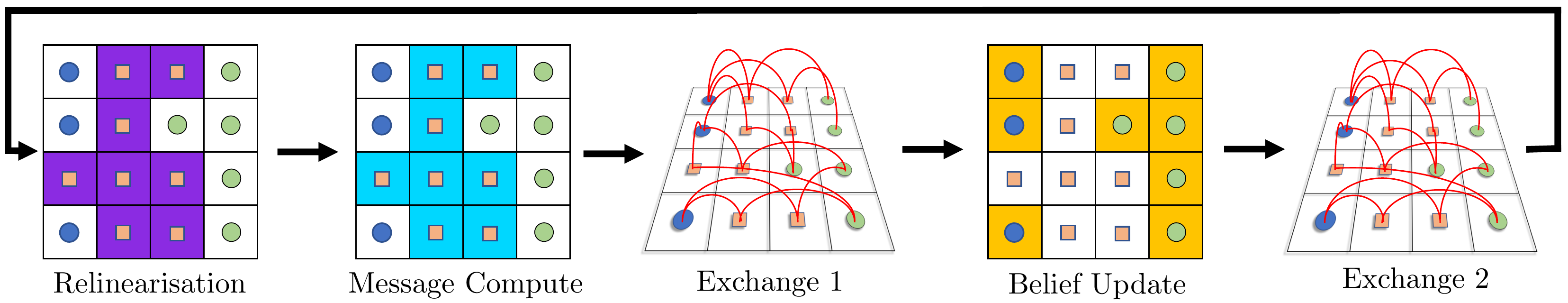}
    \vspace{1mm} \hrule \vspace{1mm}
    \includegraphics[width=\linewidth]{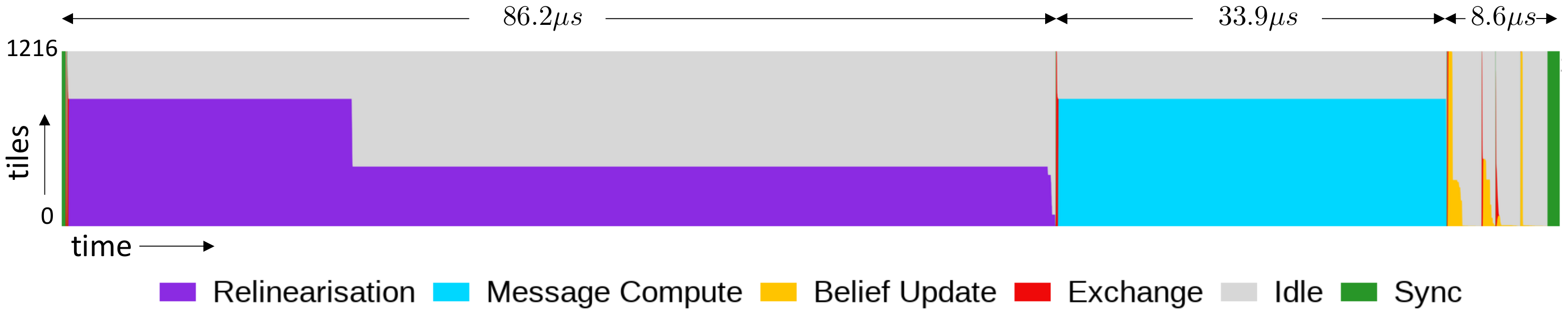}
    \caption{\textbf{IPU Phases.} \textit{Above:} A schematic showing the compute on 16 tiles in a single iteration of GBP. Tiles are coloured when they are in a compute phase. In Exchange 1, factor nodes send messages to variable nodes and in Exchange 2 variable nodes send messages to factor nodes. Keyframe and landmark variable nodes are blue and green respectively and factor nodes are orange. \textit{Below:} Plot shows the activity of each tile during a single iteration of GBP for a factor graph with 1216 nodes mapped 1-to-1 onto the tiles. In the Relinearisation phase, all 929 factors compute the distance of the adjacent beliefs from their linearisation point and a subset of these factors subsequently relinearise. The Belief Update is implemented with Graphcore's Poplibs\texttrademark library and so is significantly faster and is indicative of the speed-ups possible with a more specific implementation using an optimised linear algebra library.}
    \label{fig:ipu}
\vspace{2mm} \hrule
\end{figure*}

\section{Robust Factors}

It is well understood that measurements from real sensors usually have a distribution with gross outliers which is better represented by a function with heavier tails than a pure Gaussian measurement model. We can straightforwardly use such a robust cost function in our measurement factors within GBP.
We employ a Huber function, which transitions from the usual quadratic cost to a linear cost when the Mahalanobis distance $M_{km}(\vecx_k, \vecl_m) = \;\parallel \vecz_{km} - \vech(\vecx_k,\vecl_m) \parallel_{\matSigma_M}$ exceeds a threshold $N_\sigma$.

In order to maintain the Gaussian form of the factors in the linear loss regime, following \cite{Davison:Ortiz:ARXIV2019, Agarwal:etal:ICRA2013} we rescale the covariance of the noise in the Gaussian measurement model such that the contribution to the objective is equivalent to the Huber loss at this value. This has the effect of down-weighting or reducing the information of messages outgoing from this measurement factor. A measurement factor $\psi_{km}$ then takes the following form before linearisation \cite{Davison:Ortiz:ARXIV2019}:
\beq
    \psi_{km}(\vecx_l, \vecl_m) \propto
    \small
    \begin{cases}
    \exp{(-\frac{1}{2} M_{km}^2)}  & ,M_{km} \leq N_\sigma \\
    \exp{(-\frac{1}{2} M_{km}^2 [\frac{2 N_\sigma}{M_{km}} - \frac{N_\sigma^2}{M_{km}^2} ] )}  & ,M_{km}  \geq N_\sigma
    \end{cases}
    ~.
\eeq

\vspace{-1em}
\section{IPU Implementation}

An IPU chip is massively parallel, containing 1216 independent compute cores called {\emph{tiles}. Each tile has 256KB local memory and 6 hardware threads that can all execute independent programs. In contrast, a GPU has very limited cache on chip, all data must be fetched from off chip DRAM, and there is less flexibility for executing different programs on each thread. The IPU's distributed on-chip SRAM means that memory accesses consume approximately 1pJ per byte whereas external DRAM accesses on a GPU/CPU consume hundreds of pJ per byte. Embedded variants of the IPU will therefore have significant power advantages over existing processors \cite{Graphcore}.

To implement GBP on the IPU we must map each node in the factor graph onto a tile on the IPU. The tiles are connected all-to-all with similar latency between all pairs of tiles on a chip \cite{Jia:etal:ARXIV2019} meaning that nodes can be mapped to arbitrary tiles. The most simple mapping places exactly one factor or variable node per tile, as in Figure \ref{fig:mapping}, but limits the size of the factor graph to 1216 nodes. Noting that variable and factor nodes alternate in compute and that there are 6 threads per tile, in all experiments we are able to map much larger graphs to a single chip by placing multiple nodes per tile without affecting speed.

In order to exploit this parallelism the IPU employs a \textit{bulk synchronous parallel} execution model. In this model all tiles compute in parallel using their local memories. When each tile has finished computing it enters a waiting phase (idle). When all tiles are finished, there is a short synchronisation phase (sync) across all tiles before data is copied between tiles with extremely high bandwidth in a predetermined schedule (exchange). This process then repeats as all tiles re-enter the compute phase. The period between syncs is not fixed but determined by the time taken for the computation. 

GBP has three compute phases and two exchange phases in a single iteration. As shown in the upper part of Figure \ref{fig:ipu}, factor nodes first relinearise and then compute their messages which are sent to adjacent variable nodes before the variable nodes update their beliefs which are sent back to adjacent factor nodes. The lower part of Figure \ref{fig:ipu} shows that the total time for a single iteration of GBP is less than 125$\mu s$ while factor relinearisation and message compute makes up the bulk of the total compute time.

\section{GBP Implementation}

In experiments, we set the relinearisation threshold $\beta = 0.01$ and allow a factor to relinearise at most every 10 iterations. The damping is set to $d = 0.4$ and messages from factors are undamped for 8 iterations after relinearisation. This damping schedule allows newly relinearised messages to propagate through the graph while also stabilising later iterations. As the IPU handles halves and floats but not doubles, we found that it was necessary for numerical stability to use the Jacobians to automatically set prior constraints to initially have the same scale as the measurement constraints. These priors are then weakened to a hundredth of the strength gradually over 10 iterations. GBP is not sensitive to the mean of the prior and displays the same behaviour on convergence as when implemented on a CPU with doubles when the stronger priors are not required.

\begin{figure*}[]
    \centering
    \includegraphics[width=\linewidth]{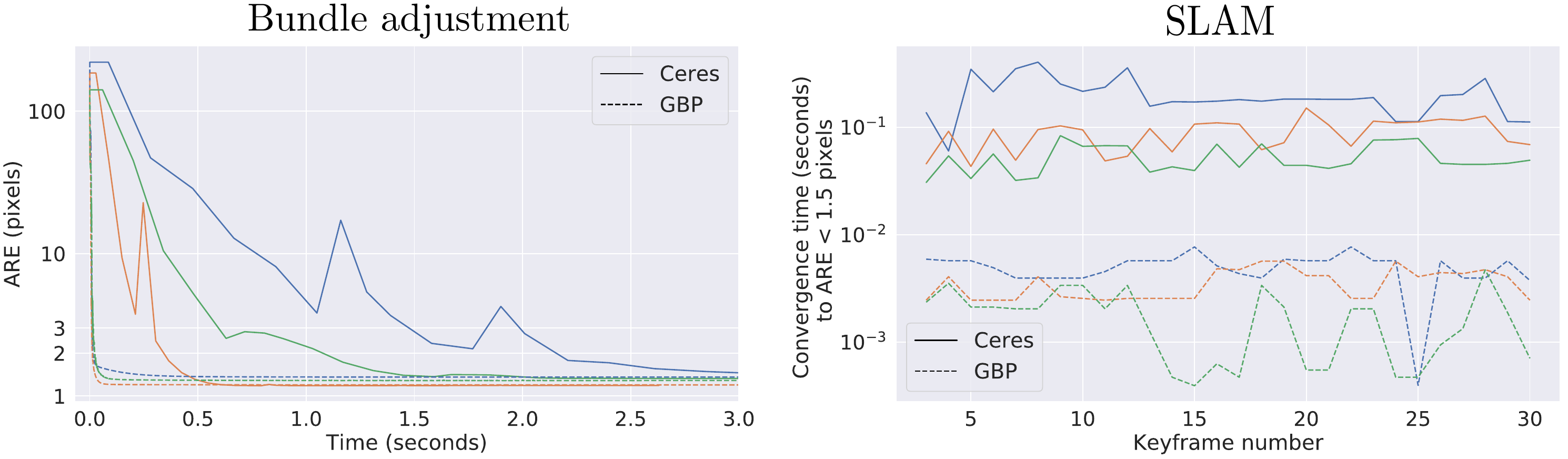}
    
    \caption{\textbf{Speed Comparison.} Note the logarithmic scale on the y axes. \textit{Left:} \textbf{ Bundle adjustment.} ARE for 3 sequences \textcolor{RoyalBlue}{fr1desk}, \textcolor{orange}{fr2desk}, \textcolor{ForestGreen}{fr3teddy}. fr1desk is more difficult as it has the most measurements and the camera moves a large distance. fr3teddy has 125 keyframes but is easier to solve as fewer landmarks are densely observed in object reconstruction. Similar results were observed for the other TUM sequences whose convergence times are described in Table \ref{tab:speed}. \textit{Right:} \textbf{SLAM. } Time to converge to ARE $<$ 1.5 pixels after a new keyframe is added and initialised with the pose of the most recent keyframe. Results are for the first 30 keyframes of the sequences \textcolor{RoyalBlue}{fr1desk}, \textcolor{orange}{fr2desk}, \textcolor{ForestGreen}{fr3teddy}. }
    \label{fig:speed_comparison}
    \vspace{2mm} \hrule
\end{figure*}

\section{Experimental Evaluation} \label{sec:results}

For evaluation we use sections of sequences from the TUM \cite{Sturm:etal:IROS2012} and KITTI \cite{Geiger:etal:CVPR2012} data sets. We use ORBSLAM \cite{Mur-Artal:etal:TRO2015} as the front-end to select keyframes, generate ORB features \cite{Rublee:etal:ICCV2011} and handle correspondence. In all TUM experiments, landmarks are initialised at a depth of 1m from the first keyframe by which they are observed, while in KITTI experiments we initialise landmarks with Gaussian noise of standard deviation 0.5m.

We compare our implementation of GBP to Ceres
\cite{CeresManual}, a non-linear least squares optimisation library often used for bundle adjustment. In all comparisons Ceres is run on a 6 core i7-8700K CPU with 18 threads (which we found experimentally to maximise performance) and uses Levenberg-Marquardt with Dense Schur and dense Cholesky on the reduced system, a Huber kernel and analytic derivatives.

\subsection{Bundle Adjustment Speed Evaluation}

First we present results to show that our implementation of GBP can rapidly solve large bundle adjustment problems. 

We evaluate the optimisation speed by tracking the average reprojection error (ARE) over all measurements in the graph. Table \ref{tab:speed} shows the time to converge to ARE $< 1.5$ pixels for 10 sequences with diverse camera motion and co-observation of landmarks in which keyframe positions are initialised with Gaussian noise of standard deviation 7cm. The corresponding ARE curves for 3 of the sequences are plotted on the left in Figure \ref{fig:speed_comparison}. GBP reaches convergence an average of 24x faster than Ceres over the 10 sequences. Typically GBP takes between 50-300 iterations to converge and Ceres takes between 10-40 steps, however, due to the rapid in-place computation on the IPU, which operates at 120W, GBP is significantly faster.

\begin{table}[]
\centering
\small
\caption{The final two columns give the time in milliseconds to converge to ARE $< 1.5$ pixels for 10 sequences from the TUM data set (two testing sequences, 4 handheld camera sequences, 2 robot mounted sequences, 2 object reconstruction sequences) and 2 from the KITTI data set. k is the number of keyframes, p landmarks, m  measurements. }
\label{tab:speed}
\begin{tabular}{| c || c c c | c c |}
    \hline
    Sequence & k & p & m & GBP & Ceres \\
    \hline
    fr1xyz & 42 & 2194 & 12908 & \textbf{37.2} & 1180 \\
    fr1rpy & 34 & 1999 & 8920 &  \textbf{130.3} & 1030 \\
    fr1desk & 63 & 2913 & 13514 & \textbf{77.3} & 2850 \\
    fr1room & 20 & 1467 & 5388 & \textbf{31.7} & 779 \\
    fr2desk & 40 & 892 & 3995 &  \textbf{20.8} & 425 \\
    fr3loh & 36 & 1140 & 5065 & \textbf{44.6} & 470 \\
    fr2robot360 & 40 & 333 & 1745 & \textbf{51.5} & 212 \\
    fr2robot2 & 20 & 567 & 4036 & \textbf{8.6} & 345 \\
    fr1plant & 40 & 1824 & 6818 & \textbf{31.8} & 1450 \\
    fr3teddy & 125 & 1919 & 9032 & \textbf{40.0} & 1450 \\
    KITTI00 & 30 & 2745 & 16304 & \textbf{14.2} & 342 \\
    KITTI08 & 30 & 3053 & 10480 & \textbf{14.8} & 394 \\
    \hline
\end{tabular}

\vspace{2mm} \hrule
\end{table}

\subsection{SLAM Speed Evaluation}

In GBP, the confidence in the belief estimations grows over iterations as the beliefs tend towards the marginal distributions. This Bayesian property is an inherent advantage over batch methods that make point estimates in the SLAM setting. For GBP, new variables are quickly snapped into a state that is consistent with the current estimates given the new constraints, while for batch methods, the full solution must be recomputed to refine just a few variables. 

We go towards validating this advantage in incremental SLAM by comparing the time taken to converge to ARE $< 1.5$ pixels after each new keyframe is added for 3 TUM sequences with 30 keyframes. New keyframes are initialised at the location of the most recent keyframe and new landmarks at a depth of 1m. To aid Ceres and mimic the Bayesian approach, we fix the landmarks for the first 3 steps of Levenberg-Marquardt optimisation. Results are shown in the right plot in Figure \ref{fig:speed_comparison} for which on average, over the 90 keyframes added, GBP converges 36x faster than Ceres, often in fewer than 10 iterations.

\subsection{Robustness Evaluation}

We compare the robustness of GBP and Ceres in solving BA problems by varying the noise added to the keyframe initialisation and counting the proportion of successful convergences over 100 trials at each noise level. Figure \ref{fig:convbasin} shows that GBP has a comparable convergence radius to Ceres for these two TUM sequences.

\bfig
    \centering
    \includegraphics[scale=0.28]{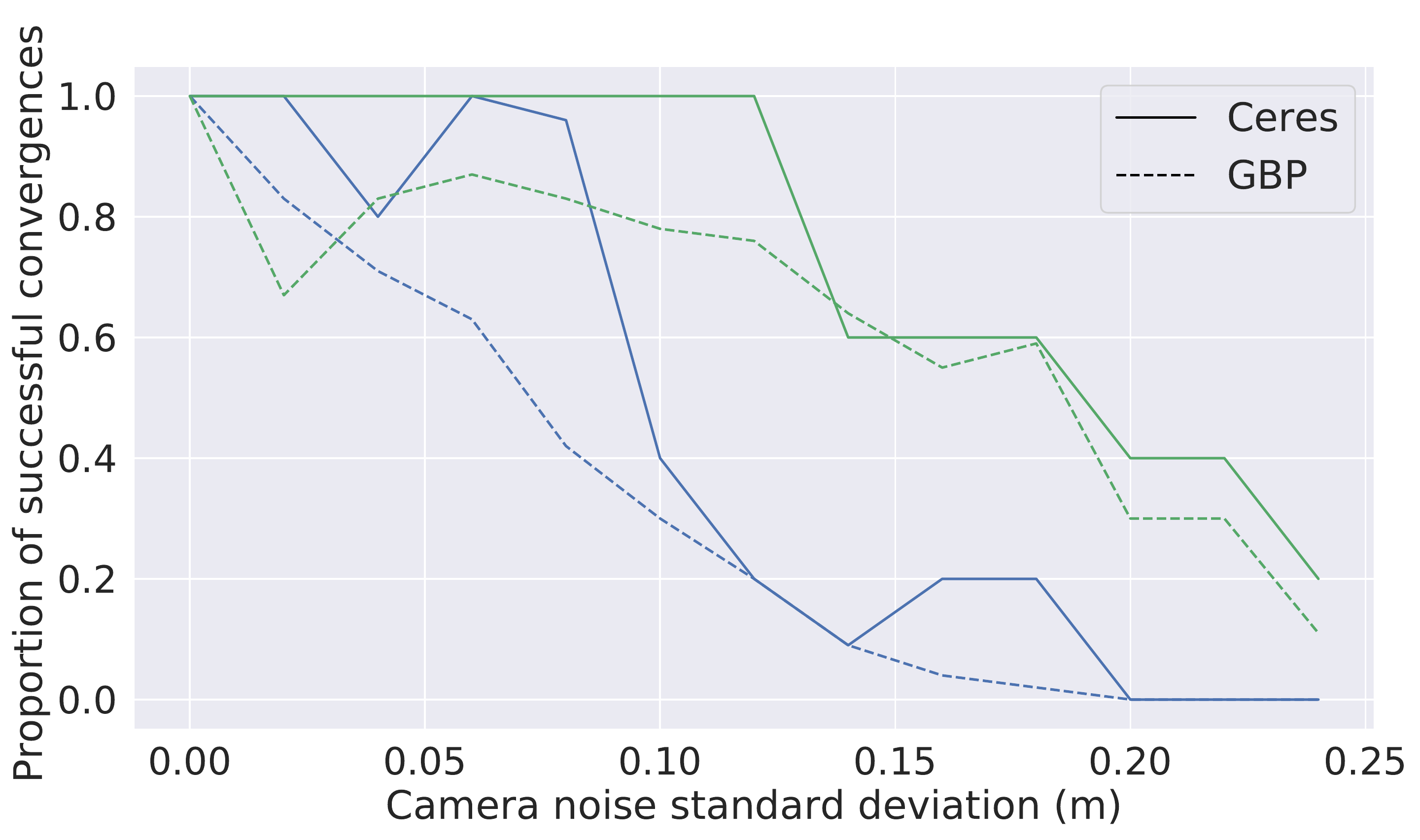}
    \caption{\textbf{Convergence basin comparison.} Proportion of successful convergences over 100 trials for different noise levels with the \textcolor{RoyalBlue}{fr1desk} and \textcolor{ForestGreen}{fr3teddy} TUM 30-keyframe sequences. A successful convergence constitutes reaching ARE $< 1.5$ pixels. }
    \label{fig:convbasin}
    \vspace{2mm} \hrule
    \vspace{-0.8em}
\efig

\subsection{Huber Loss Evaluation}

\begin{figure}[t]
    \centering
    \includegraphics[scale=0.50]{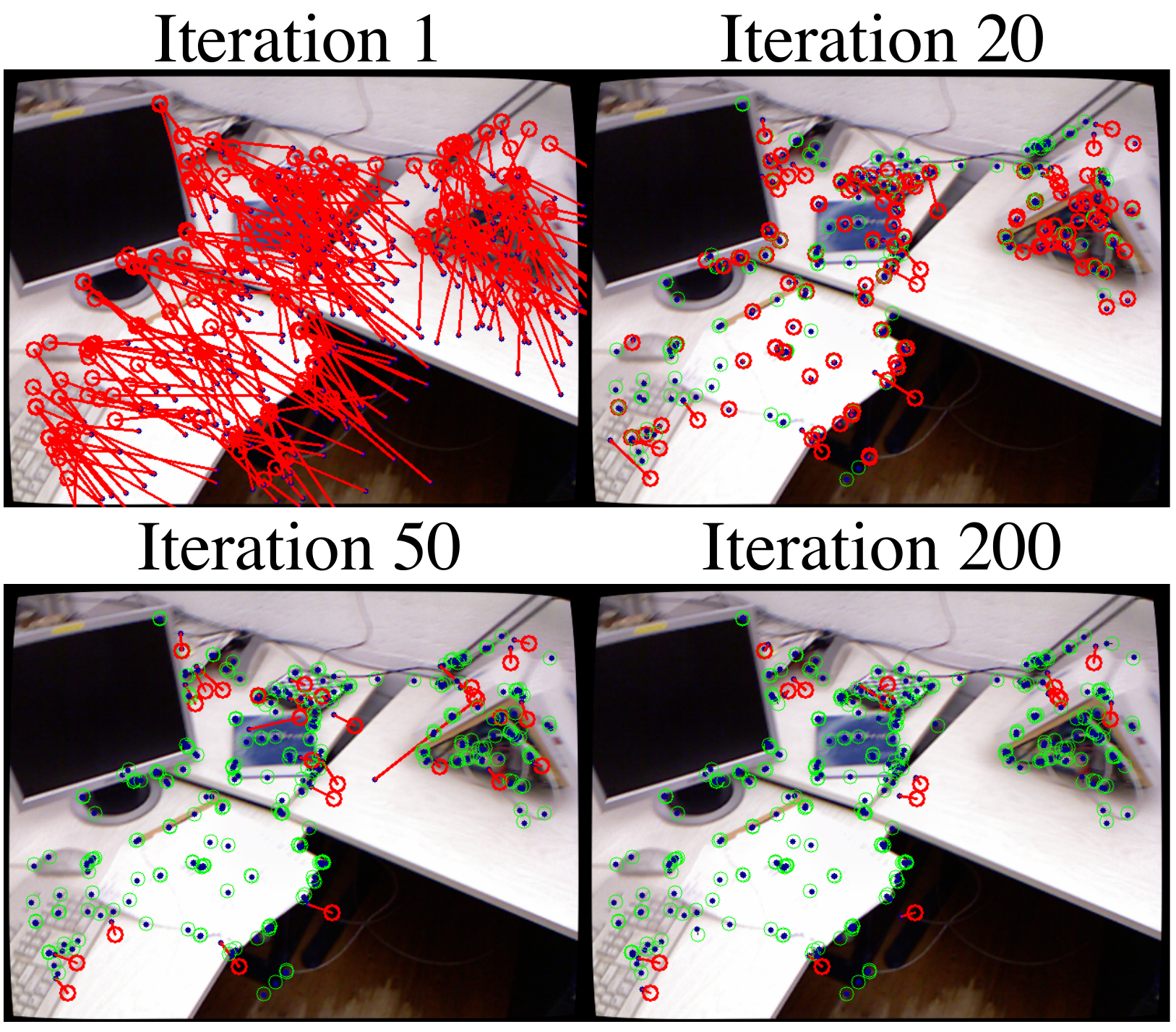}
    \caption{\textbf{GBP with Huber loss.} Landmark projections (blue points) and measurements (circles) are connected by lines. The lines and circles are red when the reprojection error exceeds the Huber threshold and the down-weighting of the message is proportional to the length of the red line. }
    \label{fig:huber_visual}
    \vspace{2mm} \hrule
    \vspace{-1em}
\end{figure}

The Huber loss function has the effect of down-weighting messages from factors that may contain outlying measurements. We demonstrate this effect in Figure \ref{fig:huber_visual} in which we visualise the reprojection errors at iterations 1, 20, 50 and 200 of GBP in a chosen keyframe for which 10\% of measurements are artificially added outliers. All measurements begin in the outlier regime and after 20 iterations a large proportion of the measurements remain in this regime as GBP has not yet worked out which measurements are inliers. By iteration 200, only the erroneous measurements are in the outlier regime as GBP has determined that these measurements are least consistent with other constraints in the graph. 
This behaviour of gradually removing false positive outlier classifications can be observed in Figure \ref{fig:huber_reproj_precision}a, for a sequence in which 3\% of data associations are incorrect. 

To validate quantitatively the benefits of the Huber loss with both GBP and Ceres, we conduct an ablation study on a sequence with incorrect data associations and measure the converged reprojection error. Figure \ref{fig:huber_reproj_precision}b shows that for GBP, the Huber loss is necessary and effective in handling incorrect data associations. For Ceres however, the same Huber loss is unable to identify the outliers and Ceres cannot arrive at a low ARE solution. This indicates that GBP's local consideration of outliers may be more effective than the global consideration in LM.

\bfig
    \centering
    \subfloat[]{{\includegraphics[scale=0.251]{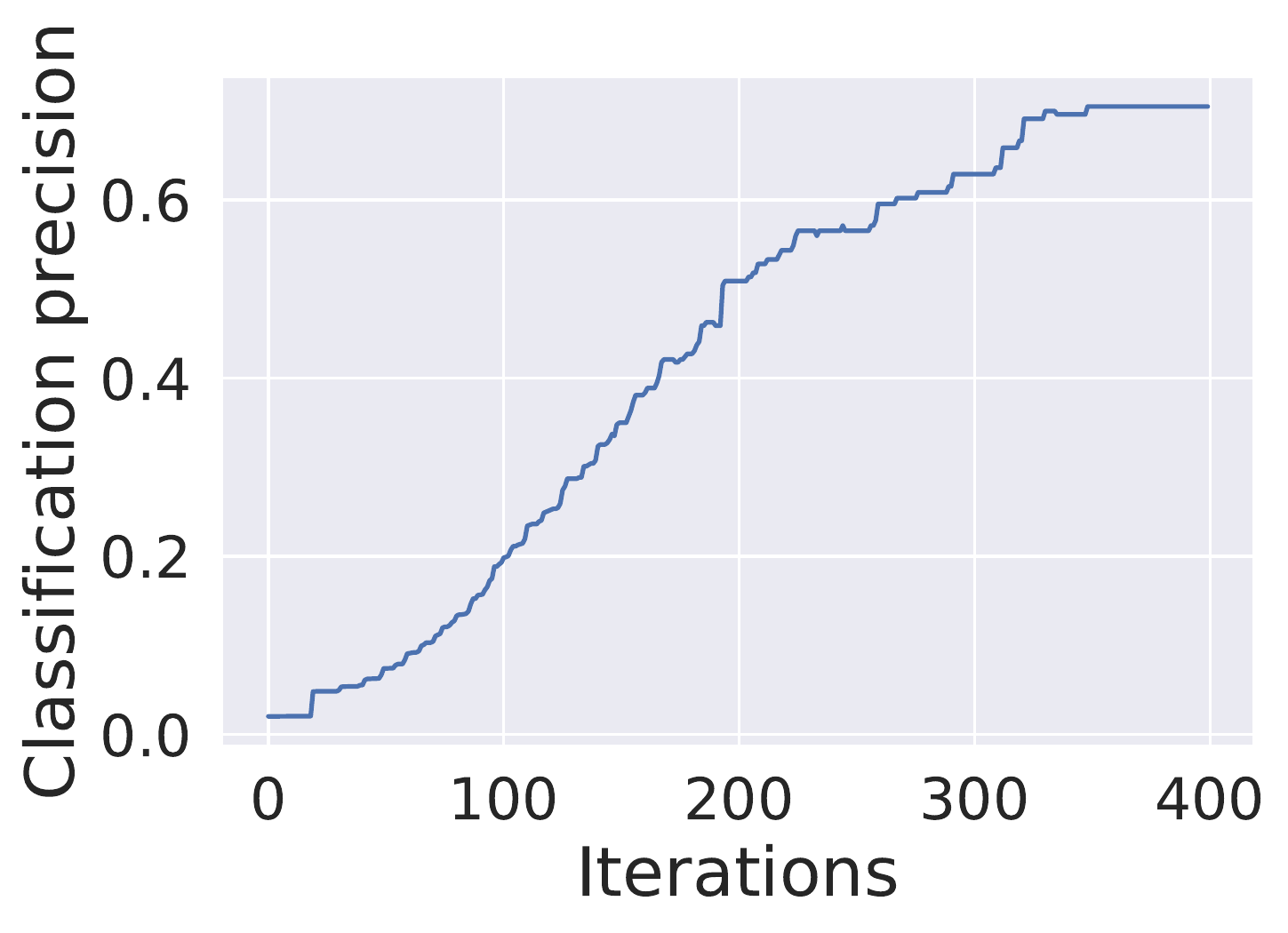} }}
    \quad
    \subfloat[]{{\includegraphics[scale=0.251]{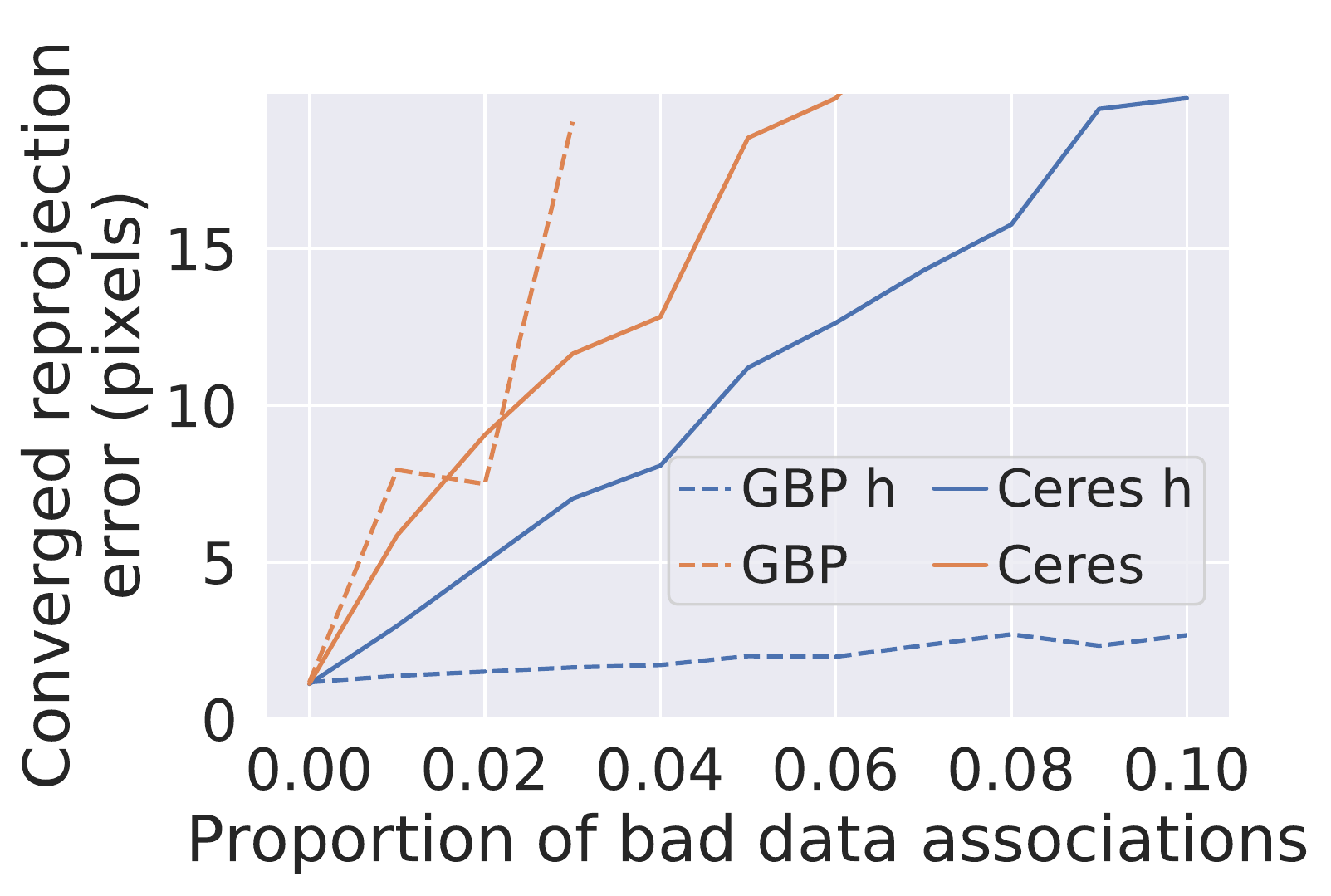} }}
    
    \caption{Results for a 20 keyframe sequence from fr1desk in which bad data associations are artificially added. \textit{(a)} Measurements are classified as outliers if they are in the linear loss regime. The recall is 1 over all iterations. ARE converges to $<$ 1.5 pixels after 268 iterations while the precision is still increasing.  \textit{(b)} h indicates Huber loss is used. For GBP, convergence is not reached without a Huber loss for more than $3\%$ bad associations, while with a Huber loss GBP can down-weight the outliers and solve the bundle adjustment problem. For Ceres, the Huber loss improves the final ARE however it still cannot converge the solution.  }
    \label{fig:huber_reproj_precision}
    \vspace{2mm} \hrule
    \vspace{-1em}
\efig

\section{Discussion / Conclusion}

We have shown that with the emergence of new flexible computer architecture for AI, specifically Graph Processors like Graphcore's IPU, Gaussian Belief Propagation can be a flexible and efficient framework for inference in Spatial AI problems. By mapping the bundle adjustment factor graph onto the tiles of a single IPU, we demonstrated that GBP can rapidly solve a variety of bundle adjustment problems with a 24x speed advantage over Ceres. Additionally, we gave an indication of the framework's capacity to efficiently solve incremental SLAM problems and be robust to outlying measurements. 

In the near term, we would like to apply GBP to very large bundle adjustment problems. Our framework scales arbitrarily to multiple chips, and Graphcore provide a custom interconnect for highly efficient inter-IPU message passing.
An even more interesting direction which looks towards low power embedded Spatial AI would investigate how to fit large problems on a single chip by merging or replacing factors using a combination of network priors and marginalisation. 
We hope that our framework of flexible, in-place optimisation on a dynamically changing factor graph will be applied to a broad spectrum of AI tasks incorporating heterogeneous factors.

\small \section*{Acknowledgements}
We thank Tristan Laidlow, Jan Czarnowski and Edgar Sucar for fruitful discussions. 

\newpage

{\small
\bibliographystyle{ieee_fullname}
\bibliography{robotvision}
}

\end{document}